\documentclass[review]{elsarticle}

\usepackage{lineno,hyperref}
\modulolinenumbers[5]

\usepackage{graphicx}
\usepackage{array}

\journal{Knowledge-Based Systems}

\begin{document}

\begin{frontmatter}

\title{Measuring pattern retention in anonymized data -- where one measure is not enough}%\tnoteref{mytitlenote}}

%\author{asdf}%\fnref{myfootnote}}
%\address{asdf}

% or include affiliations in footnotes:
\author[mymainaddress]{Sam Fletcher\corref{mycorrespondingauthor}}
\cortext[mycorrespondingauthor]{Corresponding author}
\ead{safletcher@csu.edu.au}

\author[mymainaddress]{Md Zahidul Islam}
\ead{zislam@csu.edu.au}

\address[mymainaddress]{Centre for Applied Machine Learning (CAML), School of Computing and Mathematics, Charles Sturt University, Bathurst 2795, Australia}

\begin{abstract}
In this paper, we explore how modifying data to preserve privacy affects the quality of the patterns discoverable in the data. For any analysis of modified data to be worth doing, the data must be as close to the original as possible. Therein lies a problem -- how does one make sure that modified data still contains the information it had before modification? This question is not the same as asking if an accurate classifier can be built from the modified data. Often in the literature, the prediction accuracy of a classifier made from modified (anonymized) data is used as evidence that the data is similar to the original. We demonstrate that this is not the case, and we propose a new methodology for measuring the retention of the patterns that existed in the original data. We then use our methodology to design three measures that can be easily implemented, each measuring aspects of the data that no pre-existing techniques can measure. These measures do not negate the usefulness of prediction accuracy or other measures -- they are complementary to them, and support our argument that one measure is almost never enough.
\end{abstract}


\end{frontmatter}

%\linenumbers

\section{Introduction}
\label{sec:intro}

When data contains information about people, preserving the privacy of those people is often an important concern. For example, government legislation might require a minimum level of anonymization (i.e. de-identification) of any data accessible to parties who were not given explicit permission by the individuals the data describes. Alternatively, individuals may refuse to provide their data if strong privacy guarantees are not made. Of course, the point of collecting the data in the first place is usually to discover interesting and useful patterns, and so the preservation of privacy needs to be done in a way that also preserves the utility of the data. While discovering patterns can be done manually using expert domain knowledge, the size and complexity of modern datasets has led to increasing reliance on data mining techniques. Modern datasets that contain information about people include medical data, financial data, social data, and law-enforcement data, among others. They are also often very large, sometimes containing data about hundreds of thousands of individuals. Data mining techniques such as decision forests \cite{Breiman2001,Islam2011a}, association rule mining \cite{Evfimievski2004} and frequent pattern mining \cite{Han2007} are applied to these datasets in order to extract patterns, where the patterns are usually in the form $X \rightarrow y$. $X$ is a collection of antecedents (i.e. conditions or requirements) that when true for a record $r$ (i.e. instance or tuple), predict that a consequent (i.e. label or class value) $y$ is also true for that record. Usually, each record is a collection of attributes $A$ (i.e. features) describing an individual person, and the antecedents in $X$ will contain conditions for specific values $v\in a$ of attributes $a\in A$ that some records will meet and others will not.

In order to preserve privacy, the data can be modified in a variety of ways depending on the how the data is accessible to parties beyond the data owners (henceforth referred to as ``the public''). If the data owners are fully releasing the data to the public after some modification, noise might be added to each individual's values in a way that maintains the overall distribution of values while hiding the original values of any single individual. Common techniques include \emph{additive noise} \cite{Agrawal2000,Agrawal2001,Islam2011} and \emph{multiplicative noise} \cite{Liu2006}. Alternatively, groups of values could be ``generalized'' to a single value, making values that were once different indistinguishable from each other. This is the approach \emph{k-anonymity} \cite{Sweeney2002a} and its sibling techniques (e.g. \emph{l-diversity} \cite{Machanavajjhala2007}) use. If the data is remaining under the control of the data owner and the public is merely allowed to query the data, output perturbation can be used to modify the results of the individual queries. Differential privacy \cite{Dwork2006,Dwork2006a,McSherry2007} is the most well-known technique to use this approach. Differential Privacy can also be used to generate a synthetic dataset, where new records are created using information from the original dataset. Henceforth in the paper we will be referring to ``modified'' versions of the original data, but synthetic data is an equally valid application of our proposal. Regardless of the method used to modify the data, the aim is the same: to protect the privacy of each individual in the data, without destroying the utility of the data.

Some degradation of dataset $D$'s utility is unavoidable though, since the data is no longer the same after anonymization has occurred -- the data is less truthful by definition. This is known as the privacy-utility trade-off, and optimizing this trade-off is key to a successful privacy-preservation technique. In order to assess the utility of a dataset modified to preserve privacy, it is currently common practice \cite{Brickell2008,Fung2005,Islam2011} to use a variety of data mining techniques (such as decision forests) to discover patterns in the modified dataset $M$, and then see if those patterns can correctly predict the labels of future records. ``Future records'' are simulated by excluding some (unmodified) records from the data mining process. Since the labels of these excluded records are already known, the user can tell if the patterns discovered in $M$ predicted the labels correctly. Records used in this way are often known as the ``test dataset'', $T$ \cite{Han2006a}. The Prediction Accuracy of the patterns discovered in $M$ can then be compared to the Prediction Accuracy of the patterns discovered in the original dataset $D$. Other measures such as F-measure \cite{Rijsbergen1979} and AUC \cite{Hanley1982} can be similarly used.

However this approach of measuring utility has a shortcoming: it cannot tell the user if the patterns discovered in $M$ are the same patterns discovered in $D$. There is no way of knowing if the modifications made to $D$ caused the original patterns to change (or disappear), or if weaker patterns became strong enough to become more prominent. We  propose a solution to this shortcoming of the traditional approach: a methodology for measuring the retention of the original patterns in $M$. We implement our proposed methodology with three straightforward measures of pattern retention, but they are by no means exhaustive. What makes a pattern useful to a user can vary wildly depending on their needs, and it would be misguided to blindly apply a ``one size fits all'' measure to all privacy preservation scenarios, devoid of context.

\begin{table}%[!th]
\centering
\caption{A selection of patterns discovered in the ``Adult'' dataset \cite{Bache2013}.}
\label{table:patterns}
\scriptsize
	\begin{tabular}{>{\centering\arraybackslash}m{0.3cm} >{\centering\arraybackslash}m{6.1cm} >{\centering\arraybackslash}m{3.9cm}}
	\noalign{\smallskip}\hline\noalign{\smallskip}	
	\textbf{$i$} & \textbf{Antecedent, $X_i$} & \textbf{Consequent, $Pr(Y=y)$} \\
	\noalign{\smallskip}\hline\noalign{\smallskip}
	0 & \raggedright{$206134 > Census~Weighting \leq 346177$ AND $Capital~Gains \leq 4737$ AND $Age > 40.5$ AND $Capital~Loss \leq 1836.5$ AND $Hours~per~Week > 52.5$} & $Pr(Income\leq\$50,000)=0.57$,\newline $Pr(Income>\$50,000)=0.43$ \\ 
	\noalign{\smallskip}\hline\noalign{\smallskip}
	1 & \raggedright{$244440.5 < Census~Weighting \leq 249542$ AND $Capital~Gains \leq 4737$ AND $Age \leq 40.5$} & $Pr(Income\leq\$50,000)=0.92$,\newline $Pr(Income>\$50,000)=0.08)$ \\
	\noalign{\smallskip}\hline\noalign{\smallskip}
	2 & \raggedright{$Census~Weighting \leq 206134$ AND $Capital~Gains>5316.5$} & $Pr(Income\leq\$50,000)=0.05$,\newline $Pr(Income>\$50,000)=0.95$ \\
	\noalign{\smallskip}\hline\noalign{\smallskip}
	3 & \raggedright{$116388 > Census~Weighting \leq 200855.5$ AND $Capital~Gains \leq 5316.5$ AND $Capital~Loss \leq 1198$ AND $Hours~per~Week \leq 53.5$} & $Pr(Income\leq\$50,000)=0.81$,\newline $Pr(Income>\$50,000)=0.19$ \\ 
	\noalign{\smallskip}\hline
  \end{tabular}
\end{table}

\subsection{Problem Statement}
\label{subsec:problem}

In this paper, we frame the problem from the perspective of the data owner, where the data owner is wishing to release their data to the public in a way that protects the privacy of each individual in the data. They do not wish to secure and maintain a server that outputs perturbed results to queries asked by the public, nor do they want to define groupings of values for each of the attributes present in the data. In this paper we will focus on the scenario in which noise is added to the values of the individuals. We use both continuous (i.e. numerical) and discrete (i.e. categorical) data. Note that the specific method of anonymization is not the focus, rather the focus is on how to measure the utility of the data once the anonymization method has been used.

The problem the data owner is facing is how to know how much degradation is occurring in when privacy-preservation techniques are applied to their data. They can use data mining techniques to find patterns in the modified data $M$ and see if those patterns are accurate, but there is no way to tell if the patterns are the same as the patterns that could be found in the original data $D$.

Which patterns the data owner deems important enough to monitor is outside the scope of this paper. What makes a pattern valuable can vary depending on the needs of the user, and measures have been developed to assess different aspects of patterns, such as a pattern's support or coverage \cite{Tan2002,Webb2002}, confidence \cite{Dasseni2001}, conciseness \cite{Padmanabhan2000}, peculiarity \cite{Ohshima2003}, or many other aspects depending on the user's needs \cite{Geng2006,Tan2004,Vaillant2004,Fletcher2015}. These measures are often collectively referred to as interestingness measures. How the patterns are discovered is also outside of the scope of this paper -- any patterns in the form $X\rightarrow y$ are applicable to the solution proposed in this paper. In our experiments, we arbitrarily use the CART decision tree algorithm \cite{Breiman1984} to generate a collection of patterns. The number of patterns that are monitored can be as high as the data owners likes.

We define a dataset $D$ as a two-dimensional table made up of independent rows, each defined by the values it possesses in each column. Each row represents a record $r\in D$, and each column represents an attribute $a\in A$. Each $a$ is made up of its own set of values, with each $r$ possessing one value $a_v$ per $a$, written as $r_a=a_v; \forall a\in A$. When $D$ is modified to preserve privacy, we denote the modified dataset as $M$. A dataset made up of records excluded from $D$ for testing purposes is symbolized as $T$.

A ``pattern'' can be defined as a rule $X \rightarrow y\in Y$, where $X$ is an antecedent, representing a set of conditions in the form $a=a_v$ that when met, predicts that a consequent $Y$ will equal $y$ \cite{Han2006a}. $Y$ is an attribute that has been chosen as the consequent (i.e. class attribute or label) of the pattern. If a record $r$ meets every condition in $X$ (i.e. $r \in \sigma_{X}(D)$)\footnote{Read $\sigma$ as the mathematical symbol for \emph{selection}. For example, $\sigma_p(q)$ is the subset of elements in $q$ for which $p$ is true. $p$ can either be a statement such as $Y=y$, or a set of statements such as $X$, in which case all statements in set $p$ must be true for an element in $q$ in order for that element to be in the set $\sigma_p(q)$.}, it is predicted to have a label $Y=y$ either with certainty or with some probability (i.e. $0 \leq P(r_Y=y) \leq 1$). Both $a$ and $Y$ can be either continuous or discrete; if continuous, other operators can be used such as $a>a_v$ and $Y<y$. Note that we abuse notation and simplify $X \rightarrow y$ to just $X$ when it is clear from context that we mean the whole pattern.

We refer to the set of all patterns $\{X_i; \forall i\}$ discovered in a dataset $D$ as $Z_D$. If $Z_D$ is used to predict $Y$ for all records in $T$, we write the average accuracy of these predictions (that is, the ``Prediction Accuracy'') as $\alpha(Z_D|T)$. In words, this can be read as ``The accuracy of $Z_D$ at correctly predicting class values of records in $T$''. Some examples of patterns can be seen in Table \ref{table:patterns}, including the probability of $r$ having a label $y$. Our proposed methodology and its implementations are independent of methods for discovering patterns -- any patterns in the form $X\rightarrow y$ are applicable, regardless of whether they were manually found, found with a decision tree or via association rule mining or frequent pattern mining, or any other method.

\subsection{Our Contribution}
\label{subsec:contribution}

Our contribution can be summarized as follows:

\begin{itemize}
	\item We propose a novel methodology for measuring the pattern retention of a dataset after it has been modified (or had synthetic data generated from it) with a privacy-preservation technique.
	\item We implement and test three measures that use our methodology and demonstrate their sensitivity to changes in pattern retention.
	\item Using a thought experiment, we demonstrate that other pre-existing measures are not sensitive to changes in pattern retention, while our measures are.
	\item We also provide a correlation matrix of our three measures and three pre-existing measures, showing that there is almost no correlation between the performance of a classifier built from the modified data, and the retention of the original patterns. This demonstrates that no single measure can be expected to inform the user (e.g. the data owner) about every change in the data after anonymization, and that our methodology captures information that no pre-existing measure does.
\end{itemize}

We also make the code for our three implementations of our proposed methodology available online.\footnote{The code can be found at \texttt{http://csusap.csu.edu.au/{\raise.17ex\hbox{$\scriptstyle\mathtt{\sim}$}}zislam/} or you can email us.}

In Sect.~\ref{sec:main} we propose a generalized methodology for measuring the retention of patterns in modified data. In Sect.~\ref{sec:implementations} we present three implementations of the proposed methodology. In Sect.~\ref{sec:relatedwork} we discuss three pre-existing utility measures. In Sect.~\ref{sec:thought} we use a thought experiment to explore our measures alongside pre-existing measures. In Sect.~\ref{sec:method} we detail our experiments, and in Sect.~\ref{sec:results} we present our empirical findings. We summarize our thoughts and conclude the paper in Sect.~\ref{sec:conclusion}.

\section{A Methodology for Measuring Pattern Retention}
\label{sec:main}

\begin{figure}[!b]
\centering  
\includegraphics[width=1.0\textwidth]{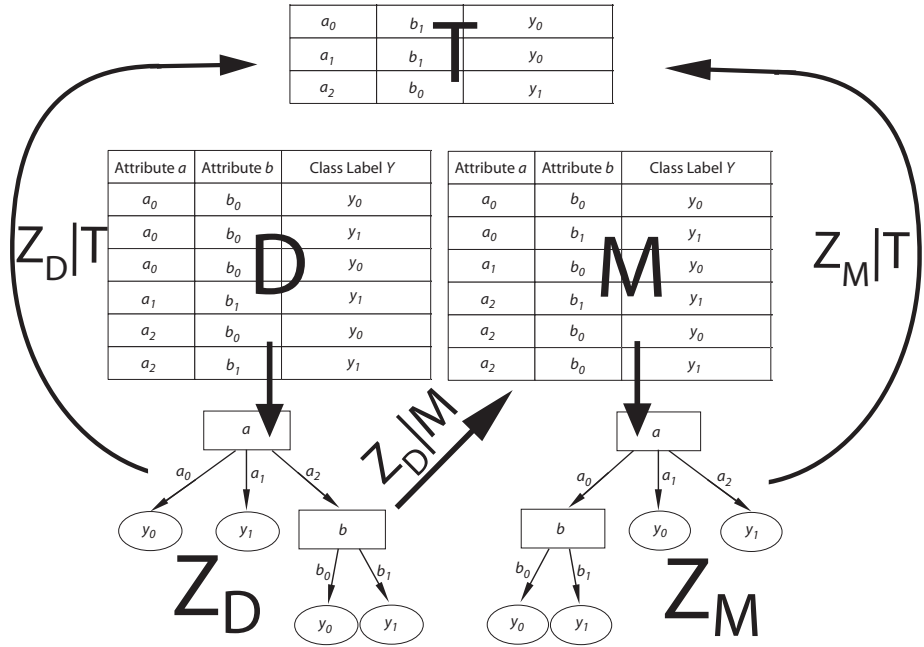}
\caption{A diagram of the data and classifiers involved when modifying data and testing for changes in quality. $T$ is drawn from the same distribution that $D$ was drawn from, without replacement. $M$ is a modified version of the data found in $D$. $Z_D$ is any kind of classifier built using $D$, and similarly for $Z_M$ and $M$. $Z_D|M$ notates an assessment of $Z_D$ when $M$ is inputted into it (and similarly for other classifiers and inputs).}
\label{fig:OP_diagram}
\end{figure}

While Prediction Accuracy is an excellent measure when evaluating the utility of a classifier or model \cite{Ferri2009,Sokolova2009}, care needs to be taken when extending its use to the privacy-preservation domain. It has been common in the past for privacy-preservation techniques to have their effect on the quality (utility) of the data measured with prediction accuracy \cite{Brickell2008,Fletcher2014,Fung2005,Islam2011}. This necessitates applying a data mining technique to the anonymized data $M$ to build a classifier, or discovering a collection of patterns with another technique, and then testing the ability for that collection of patterns\footnote{Note that a classifier is semantically the same thing as a collection of patterns if it can be broken down into antecedents and consequents.} $Z_M$ to accurately predict the class value of records in a testing dataset $T$. The accuracy of $Z_M$ can then be compared to the accuracy of $Z_D$ (i.e. $\alpha(Z_D|T) - \alpha(Z_M|T)$), and the difference is considered to be how much the privacy-preservation technique has affected the data. See Fig. \ref{fig:OP_diagram} for a graphical representation of the data and classifiers used in this discussion. 

We see two problems with this current methodology:
\begin{enumerate}[(1)]
\item It only tells the user if the particular technique used to find the patterns in $M$ produced a good classifier / model / list of patterns. Perhaps some amount of implicit assumptions can be made about the ability of other techniques to perform similarly well (``technique $f$ did well, so techniques $g$ and $h$ probably produce similar results''), but there is by no means anything explicitly said by one $Z_M$ about the universal quality of all data mining techniques applied to $M$.
\item The user cannot tell if the patterns in $Z_M$ are the same patterns that can be discovered in $D$ (such as the patterns that would be discovered if the same data mining technique was applied to $D$, producing $Z_D$). Some patterns might still be there, while others might not, and the patterns that are still there might have changed in any number of ways (such as changes in the support or confidence of the pattern, or the values $a_v$ used in the conditions in $X$).
\end{enumerate}

A solution to (1) is for the data owner to build a $Z_M$ and $Z_D$ with every possible data mining technique they think is worth checking \cite{Brickell2008}. This solution is extremely computationally expensive, and does not address (2). If this solution is not used, then a user must either trust the implicit assumption that other data mining techniques will perform similarly, or release the collection of patterns $Z_M$ that they \emph{did} test, and not release $M$ to the public at all \cite{Brickell2008}. To the best of our knowledge, aside from our preliminary investigation \cite{Fletcher2014}, no solution to (2) currently exists in the literature.\footnote{This paper is an extension of that investigation, not previously published in a journal.}

We therefore propose a methodology that addresses both problems. Rather than discovering a collection of patterns in $M$ (i.e. $Z_M$) that may or may not have any relation to the patterns in $D$ (i.e. $Z_D$), we propose that the data owner defines a collection of patterns found in $D$ and evaluates if the data in $M$ still follows those patterns. No $Z_M$ needs to be computed in order to measure the pattern retention of $M$. Nor is $T$ required, unless it is desired for other, unrelated testing.

This methodology removes the problem described in (1), since there is no longer any data mining technique being applied to $M$. It also solves (2) since the same patterns that were found in $D$ are being used to evaluate $M$.

The next natural question is: how exactly do we evaluate $M$ with $Z_D$? There are many potential implementations of our proposed methodology, but we provide three examples below in Sect. \ref{sec:implementations}. The first measure (in Sect. \ref{subsec:pattern_accuracy}) evaluates how much of $Z_D$ as a whole has been retained in $M$. The next two measures (in Sect. \ref{subsec:psd} and Sect. \ref{subsec:pld}) evaluate the presence of each pattern in $Z_D$ (i.e. $X_i;\forall i$) separately, offering the data owner the ability to check for changes in individual patterns, as well as seeing the average change. Other implementations can easily be designed to meet the needs of the data owner. Every dataset has its own nuances, and it is usually advantageous to take those nuances into account when measuring the effect of privacy-preservation techniques, rather than trying to use a ``catch-all'' approach. The release of data to the public will be a one-time event (once it's out there, there's no taking it back!), and so spending additional resources to properly evaluate $D$ and $M$ is likely worth it.

\section{Implementations of the Methodology}
\label{sec:implementations}

\subsection{Pattern Accuracy}
\label{subsec:pattern_accuracy}

Introduced by us in a 2014 conference \cite{Fletcher2014} and not published in a journal until now, Pattern Accuracy is a simple measure that compares $D$ and $M$. Like its name might suggest, it is very similar to Prediction Accuracy in that it measures the average accuracy of a collection of patterns at predicting the class value $Y$ of some data. However instead of predicting the class value of some testing data $T$, it predicts the class values of the modified data $M$. If the Prediction Accuracy of $M$ is written as $\alpha(Z_M|T)$, then the Pattern Accuracy of $M$ is written as $\alpha(Z_D|M)$. In a privacy-preservation scenario the point is to compare $M$'s performance to $D$, so Prediction Accuracy becomes $\alpha(Z_D|T) - \alpha(Z_M|T)$, and the Pattern Accuracy equivalent is therefore $\alpha(Z_D|D) - \alpha(Z_D|M)$.\footnote{Remember that $M$ is a modified version of $D$, with each record in $M$ corresponding to an unaltered version in $D$} Note that while $\alpha(Z_D|D)$ should not be used to assess the quality of a classifier due to the risk of of overfitting, in this paper we are not concerned with the generality of the patterns. How $Z_D$ is created or defined is outside the scope of this paper. Instead, the difference between $\alpha(Z_D|D)$ and $\alpha(Z_D|M)$ tells us the difference in the number of records that are contributing to the prediction made by each pattern (where the prediction is the majority class label). Pattern Accuracy is a way of measuring the presence of $D$'s patterns in $M$; if $\alpha(Z_D|D) - \alpha(Z_D|M)$ is close to zero, then the user knows that a similar number of records in $D$ and $M$ are contributing to the correct predictions made by the classifier built from $D$. Since the records in $M$ are just modified versions of the records in $D$, this is a valuable thing to know! If $\alpha(Z_D|D) - \alpha(Z_D|M)$ is closer to one, the user knows that the records were modified in a way that reduced the presence of the patterns found in $Z_D$. If $\alpha(Z_D|D) - \alpha(Z_D|M)$ is negative, this is actually just as bad as positive result of similar magnitude, because $D$ is trusted data -- any random modifications made to $D$ is further from the trusted data by definition, even if some quality metrics increase. Ideally we want every pattern in $Z_D$ to be just as prevalent in $M$ as it is in $D$; no more, no less. Thus we define Pattern Accuracy as:

\begin{equation}\label{eq:patt_acc}
\mbox{Pattern Accuracy} = |\alpha(Z_D|D) - \alpha(Z_D|M)| \enspace .
\end{equation}

Pattern Accuracy evaluates whether $Z_D$, as a whole, can correctly predict $Y$ for records in $M$. Since it uses an identical process to Prediction Accuracy (with the user simply having to redirect the measure to check $M$ rather than $T$), it gains all of the benefits of Prediction Accuracy such as low computation time and conceptual simplicity. What it does not do, however, is evaluate the presence of each pattern individually (i.e. $X_i;\forall i$). There are almost always multiple patterns that predict the same $y\in Y$, so it is possible that some patterns no longer have records in $M$ that follow them (and instead those records follow different patterns) without the Pattern Accuracy result changing. As long as the record's new pattern still correctly predicts $y$, the Pattern Accuracy measure is insensitive to this change. The following two measures avoid this insensitivity by evaluating the pattern retention in $M$ on a per-pattern basis, rather than evaluating the entire pattern list as a whole.

\subsection{Pattern Support Distance (PSD)}
\label{subsec:psd}

The ``support'' of a pattern is the number of records in a dataset that a pattern covers \cite{Webb2002,Dasseni2001}, and can be represented as $|\sigma_{X}(D)|$ when describing the support of pattern $X$ in dataset $D$. Whether $Y$ is predicted correctly is irrelevant when measuring support. To measure the support for $X$ in $D$ compared to $M$, we can calculate $|\sigma_{X}(D)|$ and $|\sigma_{X}(M)|$. By comparing these results, a user knows how much the presence of an individual pattern $X$ has changed due to the modifications made to $D$ (resulting in $M$). This level of granularity allows the user to use their domain knowledge to make specific assessments of the status of each $X$. This can naturally be repeated for all $X\in Z_D$. To summarize the overall support retention of $Z_D$ for a dataset $M$, the mean difference can be calculated:

\begin{equation}\label{eq:PSD}
PSD = \frac{1}{|Z_D|\times |D|} \sum_{X\in Z_D} |~|\sigma_{X}(D)|-|\sigma_{X}(M)|~| \enspace .
\end{equation}

Note that each pattern contributes equally to the mean difference. Patterns with higher support are not assumed to be more important, since each $X$ in $Z_D$ should have already been assessed by the user as being important enough to worry about preserving in $M$, and we are now only interested in if the original support has \emph{changed}. It should be noted though that patterns in $Z_D$ with very low support cannot reduce in size by as much as patterns with high support -- support cannot go below 0 -- so the presence of many patterns with low support risks ``diluting'' PSD.\footnote{This effect is caused whenever many small differences are averaged alongside several large differences. The presence of near-zero numbers effectively reduces the average, diluting the larger differences. There is nothing inherently wrong with this, but it is usually undesirable.} However it is normal for such small patterns to be considered as idiosyncrasies of $D$, and not generalizing to future records (such as $T$), and so most data mining algorithms automatically remove them from $Z_D$ \cite{Han2006a}. This is sometimes referred to as the ``minimum support threshold''.

PSD (Pattern Support Distance) has a defined lower and upper limit of $0\leq PSD\leq 1$, allowing for an intuitive interpretation of the result, such as: ``The average percentage change in the prevalence of a pattern when modifying $D$ into $M$''. 

The aim of privacy preservation is to (1) make any individual record difficult to identify, while (2) leaving the patterns as unaffected as possible \cite{Agrawal2000,Estivil-Castro1999,Islam2011}. The pattern support difference (PSD) of $M$ accurately measures the second half of this aim, allowing researchers to make more informed assessments of the overall success of a privacy preservation technique. The specific records that matched each $X$ in $D$ is irrelevant -- $X$ will still be just as prevalent in $M$ as it was in $D$ if other records take the place of the records that no longer follow $X$. In order for a record to change which pattern it matches, its values must have changed during the modification process enough for it to legitimately meet the conditions of a different pattern.

While Pattern Accuracy somewhat measures the presence (support) of $X_i\in Z_D;\forall i$ in $M$, PSD does so with precision, removing any uncertainty about the presence of each pattern in $M$.

\subsection{Pattern Label Distance (PLD)}
\label{subsec:pld}

Say a record $r\in D$ meets the conditions of a certain pattern $X_i\in Z_D$ (i.e. $r\in \sigma_{X_i}(D)$). When $D$ is modified to $M$, it is possible that $r$ will be changed in a way that causes it to meet the conditions of a different pattern $X_{j}\in Z_D$ (i.e. $r\in \sigma_{X_{j}}(M)$). If this occurs, the distribution of labels (i.e. $Y$) will change for both $X_i$ and $X_j$, since $r_Y$ has been removed from $X_i$'s distribution of class labels and added to $X_j$'s. The purpose of a pattern is often to predict $Y$, and so it is important to know how much that prediction might have changed in $M$. Pattern Accuracy measures this to an extent, as $X_i$ and $X_j$ might predict different class labels and a maximum of one of those predictions can be correct for a record $r$. But it is also possible that the two patterns will predict the same class label, leading to no change in the Pattern Accuracy of $M$ compared to $D$ (at least as far as $r$ is concerned). The consequent of any pattern $X$ is usually the most common class label to occur out of all the records in $\sigma_{X}(D)$, with any other class labels being ignored \cite{Han2006a}. This has the effect of making $X$'s prediction of $Y=y$ appear identically confident\footnote{``Confidence'' refers to the certainty or reliability of a pattern -- that is, how frequent the most frequent label is \cite{Tan2002}. If 100\% of the records in a pattern have the same class label, then that pattern can be considered highly reliable.} regardless of how high or low the frequency of $y$ is in $\sigma_{X}(M)$ compared to $\sigma_{X}(D)$, as long as it remains the most frequent class label.

To avoid these problems, we use the Chi-squared histogram distance \cite{Pele2010} to measure differences in the distribution of $Y$ between $\sigma_{X}(D)$ and $\sigma_{X}(M)$:

\begin{equation}\label{eq:chi-squared}
\chi^2(\sigma_{X}(D), \sigma_{X}(M)) = \frac{1}{2}\sum_{y\in Y} \frac{(f^D_y - f^M_y)^2}{f^D_y + f^M_y} \enspace ,
\end{equation}
where $f^D_y$ is the relative frequency of the class label $y$ in $\sigma_{X}(D)$ (in other words, the fraction of records in $\sigma_{X}(D)$ that have $r_Y=y$), and similarly for $f^M_y$ in respect to $\sigma_{X}(M)$.

Just like with Chi-squared hypothesis testing, Chi-squared histogram distance becomes unstable if there are less than five samples. This limitation is automatically handled if a minimum support threshold for each pattern $X$ was applied when making $Z_D$; otherwise we recommend discounting any patterns that have less than five class labels (i.e. ignore patterns $X\in Z_D$ where $|\sigma_{X}(D)|<5$).

Even if the majority $y$ value in $\sigma_{X}(D)$ occurs even more frequently in $\sigma_{X}(M)$ (and thus has increased confidence), this should not be considered as an improvement unless the modification that created $M$ was aiming to improve pattern utility. In scenarios such as privacy preservation, the distribution of $Y$ for $\sigma_{X}(D)$ is considered to be the ground truth. PLD (Pattern Label Distance) successfully captures this scenario, where any distance away from $D$ is a reduction in utility by definition. The mean Chi-squared histogram distance of all patterns in $Z_D$ can then be easily calculated:

\begin{equation}\label{eq:PLD}
PLD = \frac{1}{|Z_D|} \sum_{X\in Z_D} \chi^2(\sigma_{X}(D), \sigma_{X}(M)) \enspace .
\end{equation}

It should be noted that Chi-squared histogram distance is invariant to the number of records \cite{Pele2010}, and so the support of a pattern (both in $D$ and $M$) does not affect the result. If the support of each pattern is deemed relevant by the user, $|\sigma_{X}(D)|$ can easily be taken into account as well. We do not recommend combining a pattern's support difference and label distribution distance into a single result, as the results are likely to be far more informative when separate. This is true for both single patterns and the mean results (PSD and PLD). Chi-squared histogram distance is also invariant to the number of labels, so it is not restricted to datasets or patterns with a particularly sized $Y$. This is often a concern with popular measures such as AUC \cite{Hanley1982} and F-measure \cite{Rijsbergen1979}, where non-binary class attributes need to be treated with care \cite{Felkin2007}.

\section{Related Utility Measures}
\label{sec:relatedwork}

As mentioned in Sect. \ref{sec:intro}, it is common in the literature for a privacy-preserving technique's impact on data utility to be quantified using Prediction Accuracy: that is, by comparing $\alpha(Z_M|T)$ to $\alpha(Z_D|T)$ \cite{Han2006a}. Other common measures are F-measure \cite{Rijsbergen1979} and AUC \cite{Hanley1982}, where again $Z_M$ is compared to $Z_D$ using $T$ (refer to Fig. \ref{fig:OP_diagram} for a graphical representation of these data and classifiers). We use Prediction Accuracy, F-measure and AUC in our experiments.

Formally, Prediction Accuracy can be written as:
\begin{equation}\label{eq:pred_acc}
\alpha(Z_D|T) = \frac{1}{|T|} \sum_{r\in T} \mbox{\textbf{1}}(r_Y = y_{predicted}) \enspace ,
\end{equation}
where $y_{predicted}$ is the class value $y\in Y$ that $Z_D$ predicts $r$ to have, and $r_Y$ is $r$'s actual class value. The indicator function, $\mbox{\textbf{1}}(x)$, returns 1 if $x$ is true; otherwise 0. Thus, Prediction Accuracy is the fraction of records that have their class label correctly predicted.

AUC and F-measure are most appropriate when $Y$ is binary (i.e. $|Y|=2$), and where $y_1\in Y$ is the ``important'' class label, or the class label that the user is trying to correctly predict, and $y_2\in Y$ is unimportant. These labels are often referred to as the ``positive'' and ``negative'' labels, respectively. A ``True Positive'' (TP) label is therefore a label that was correctly predicted to be positive, a ``False Positive'' (FP) is a label that was predicted to be positive but was not, and similarly for ``True Negative'' (TN) and ``False Negative'' (FN).\footnote{Using this notation, we can actually re-write Prediction Accuracy as $\frac{TP+TN}{TP+TN+FP+FN}$.}

F-measure \cite{Rijsbergen1979} can be formally written in terms of precision ($\frac{TP}{TP+FP}$) and recall ($\frac{TP}{TP+FN}$) as:
\begin{equation}\label{eq:f1}
F_\beta = (1+\beta^2)\times \frac{precision~\times~recall}{\beta^2~\times~precision + recall} \enspace ,
\end{equation}
where $\beta$ is very often equal to 1, so that recall and precision have equal weighting. We use $\beta=1$ in our experiments.

Meanwhile, AUC \cite{Hanley1982} is shorthand for ``Area under the ROC curve'', with ``ROC'' in turn being short for ``Receiver Operating Characteristic''. The ROC curve describes the trade-off between TP (benefits) and FP (costs). Often it is plotted on axes with the TP Rate ($\frac{TP}{TP+FN}$) as the $y$ axis and the FP Rate ($\frac{FP}{FP+TN}$) as the $x$ axis. Thus, AUC is the area under this curve. It represents the probability that $Z_D$ is more likely to predict a positive label as positive than to predict a negative label as positive. It has become popular in the machine learning community as of late, despite some problems it has when comparing different classifiers \cite{Hand2009,Lobo2008}.

\section{A Thought Experiment}
\label{sec:thought}

We use a thought experiment to demonstrate the sensitivity of our measures to changes in the data that are not detected by pre-existing measures. We will use the toy data and classifiers seen in Fig. \ref{fig:OP_diagram}. The patterns in $Z_D$ have been written out in Table \ref{table:thought1}, along with their support and confidence. After modifying $D$ with a privacy-preservation technique, the result is $M$ as seen in Fig. \ref{fig:OP_diagram}. The classifier $Z_M$ was then made from that modified data; we present the patterns in Table \ref{table:thought2}. We then assess the quality of $M$ using six measures: our three implementations of our proposed methodology, as well as Prediction Accuracy, AUC and F-measure. The results are tabulated in Table \ref{table:thought3}.

\begin{table}
\centering
\caption{The root-to-leaf paths from $Z_D$ in Fig. \ref{fig:OP_diagram} expressed as patterns, including confidence and support.}
\label{table:thought1}
	\begin{tabular}{c c c c}
	\noalign{\smallskip}\hline\noalign{\smallskip}	
	\textbf{$i$} & \textbf{Antecedent, $X_i$} & \textbf{Consequent, $Pr(Y=y)$} & \textbf{Support} \\
	\noalign{\smallskip}\hline\noalign{\smallskip}
	0 & $a_0$ & $Pr(y_0)=0.66$,~ $Pr(y_1)=0.34$ & 3 \\ 
	\noalign{\smallskip}\hline\noalign{\smallskip}
	1 & $a_1$ & $Pr(y_0)=0.0$,~ $Pr(y_1)=1.0$ & 1 \\ 
	\noalign{\smallskip}\hline\noalign{\smallskip}
	2 & $a_2$ AND $b_0$ & $Pr(y_0)=1.0$,~ $Pr(y_1)=0.0$ & 1 \\ 
	\noalign{\smallskip}\hline\noalign{\smallskip}
	3 & $a_2$ AND $b_1$ & $Pr(y_0)=1.0$,~ $Pr(y_1)=0.0$ & 1 \\ 
	\noalign{\smallskip}\hline
  \end{tabular}
\end{table}

\begin{table}
\centering
\caption{The root-to-leaf paths from $Z_M$ in Fig. \ref{fig:OP_diagram} expressed as patterns, including confidence and support.}
\label{table:thought2}
	\begin{tabular}{c c c c}
	\noalign{\smallskip}\hline\noalign{\smallskip}	
	\textbf{$i$} & \textbf{Antecedent, $X_i$} & \textbf{Consequent, $Pr(Y=y)$} & \textbf{Support} \\
	\noalign{\smallskip}\hline\noalign{\smallskip}
	0 & $a_0$ AND $b_0$ & $Pr(y_0)=1.0$,~ $Pr(y_1)=0.0$ & 1 \\ 
	\noalign{\smallskip}\hline\noalign{\smallskip}
	1 & $a_0$ AND $b_1$ & $Pr(y_0)=1.0$,~ $Pr(y_1)=0.0$ & 1 \\ 
	\noalign{\smallskip}\hline\noalign{\smallskip}
	2 & $a_1$ & $Pr(y_0)=1.0$,~ $Pr(y_1)=0.0$ & 1 \\ 
	\noalign{\smallskip}\hline\noalign{\smallskip}
	3 & $a_2$ & $Pr(y_0)=0.34$,~ $Pr(y_1)=0.66$ & 3 \\ 
	\noalign{\smallskip}\hline
  \end{tabular}
\end{table}

\begin{table}
\centering
\caption{The results of six measures when the two patterns seen in Table \ref{table:thought1} undergo changes so that they now resemble what is seen in Table \ref{table:thought2}.}
\label{table:thought3}
	\begin{tabular}{c c c c c c c}
	\noalign{\smallskip}\hline\noalign{\smallskip}
			& \begin{tabular}[c]{@{}c@{}}\textbf{Pattern}\\ \textbf{Accuracy}\end{tabular} & \textbf{PSD} & \textbf{PLD} & \begin{tabular}[c]{@{}c@{}}\textbf{Prediction}\\ \textbf{Accuracy}\end{tabular} & \textbf{AUC}	& \textbf{F-measure} \\
	\noalign{\smallskip}\hline\noalign{\smallskip}
	\textbf{$D$} & 0.00 & 0.00 & 0.00 & 0.67 & 0.67 & 0.80 \\ 
	\textbf{$M$} & 0.50 & 0.08 & 0.34 & 0.67 & 0.50 & 0.67 \\ 
	\textbf{Change} & \textbf{0.50} & \textbf{0.08} & \textbf{0.34} & \textbf{0.00} & \textbf{0.17} & \textbf{0.13} \\
	\noalign{\smallskip}\hline
  \end{tabular}
\end{table}

Several things have happened here. Firstly, Prediction Accuracy was completely incapable of detecting any changes in $M$ compared to $D$. It is possible that the user does not actually care that $M$ is different, and is only interested in being able to make good predictions on future data (and that is fine). However, if the user makes any assumption about the similarity between $M$ and $D$ with Prediction Accuracy, they have made a very dangerous mistake. As we can see visually in Fig. \ref{fig:OP_diagram}, $Z_M$ is radically different from $Z_D$. Due to the changes present in $M$, the patterns discovered in $M$ are very different from the patterns discovered in $D$. Our proposed methodology solves the issue of quantifying the visual intuition one has about the differences between $Z_M$ and $Z_D$. Pattern Accuracy, PSD and PLD were all able to accurately identify the differences between $D$ and $M$ that they are designed to identify: the overall retention of $Z_D$'s patterns, the changes in the patterns' support and the changes in the patterns' class label distribution, respectively.

AUC and F-measure were able to detect some changes, but it is important to recognize that these changes do not represent any connection between $Z_D$ and $Z_M$. Both measures started by calculating the true and false predictions of the positive and negative labels of $Z_D$ using $T$, and then they made similar calculations of $Z_M$ using $T$. At no point was $Z_M$ actually compared to $Z_D$, except indirectly, in much the same way that Prediction Accuracy indirectly compares them. As demonstrated by Prediction Accuracy's results in Table \ref{table:thought3} though, there is no guarantee that any of these indirect comparisons will detect any differences at all. Even if they do, how does the user use that information, except to prompt them to look at the patterns discovered with their human intuition? Pattern Accuracy, PSD and PLD offer concrete results about $M$'s retention of $D$'s patterns.

\section{Experiment Methodology}
\label{sec:method}

To empirically evaluate our three measures, we carry out the below experiments and present the results in Sect. \ref{sec:results}. For our experiments, the patterns are generated using decision trees. Note that the patterns could just have easily been manually created, generated from a different classifier, filtered using any number of interestingness measures, hand-picked from a list of generated patterns, or by any other means that outputs patterns in the form $X \rightarrow y$.

We use 17 datasets publicly available in the UCI Machine Learning Repository \cite{Bache2013}. To generate a collection of patterns for each dataset (i.e. $Z_D$), we run the CART algorithm \cite{Breiman1984}, with a minimum leaf size (i.e. minimum support threshold) of $|D|\times 0.02$ and a maximum tree depth of 12. By generating patterns in this way, we produce a set of realistic patterns for each dataset, with patterns also varying in length (i.e. the number of conditions in $X$). Another advantage of generating our patterns in this way is that the deterministic nature of CART allows for others to replicate our $Z_D$'s exactly.

Our datasets range in size from 653 to 58000 records, 6 to 62 attributes, and 2 to 18 class labels, and include both numerical and categorical attributes. The number of patterns (i.e. $|Z_D|$) ranges from 11 to 37. The details of the datasets are summarized in Table \ref{table:datasets}.

For experiments involving AUC and F-measure (e.g. Table \ref{table:correlations}, discussed later) we limit our experiments to datasets with binary labels, where these measures are known to work best.

\begin{table}[t]
\centering 
\caption{Details of the datasets used in our experiments. The columns are, in order: the number of records in $D$; the number of continuous attributes in $D$; the number discrete attributes in $D$; the number of labels (class values) for $Y$ in $D$; and the relative frequency of the most common label in $Y$.}
\footnotesize
	\begin{tabular}{>{\centering\arraybackslash}m{1.8cm} >{\centering\arraybackslash}m{1.3cm} >{\centering\arraybackslash}m{1.9cm} >{\centering\arraybackslash}m{1.9cm} >{\centering\arraybackslash}m{0.9cm} >{\centering\arraybackslash}m{1.5cm} >{\centering\arraybackslash}m{1.95cm}}
	\noalign{\smallskip}\hline\noalign{\smallskip}	
	\textbf{Name}	& \textbf{Records} & \textbf{Numerical Attributes} & \textbf{Categorical Attributes} & \textbf{Labels} & \textbf{Majority Label \%} \\
	\noalign{\smallskip}\hline\noalign{\smallskip}	
	Banknotes & 1372 & 4 & 0 & 2 & 55\% \\
	Vehicle & 846 & 18 & 0 & 4 & 26\% \\	
	RedWine & 1599 & 11 & 0 & 6 & 43\% \\	
	Spambase & 4601 & 57 & 0 & 2 & 60\% \\
	Wilt & 4839 & 5 & 0 & 2 & 95\% \\
	WallSensor & 5456 & 4 & 0 & 4 & 40\% \\
	PageBlocks & 5473 & 10 & 0 & 5 & 90\% \\	
	OptDigits & 5620 & 62 & 0 & 10 & 10\% \\	
	PenWritten & 10992 & 16 & 0 & 10 & 10\% \\	
	GammaTele & 19014 & 10 & 0 & 2 & 65\% \\	
	Shuttle & 58000 & 9 & 0 & 7 & 79\% \\
	Credit & 653 & 6 & 9 & 2 & 55\% \\
	Parkinsons & 1040 & 28 & 1 & 2 & 50\% \\
	Yeast & 1484 & 7 & 1 & 10 & 31\% \\	
	Cardio & 2126 & 21 & 1 & 3 & 78\% \\	
	Adult & 30162 & 6 & 5 & 2 & 75\% \\	
	Bank & 45211 & 7 & 9 & 2 & 88\% \\
	\noalign{\smallskip}\hline
  \end{tabular}
	\label{table:datasets}
\end{table}

\subsection{Privacy-Preservation Techniques}
\label{subsec:noise}

To simulate various modifications to a dataset, we add noise to the data in two simple ways. Each type of noise represents a different scenario respectively: where attribute and multi-attribute (i.e. multivariate) distributions are flattened (i.e. made more uniform); and where attribute distributions and most multi-attribute distributions are preserved. We simulate these scenarios using additive noise. Using these two scenarios, we explore what a user can learn from our three implementations of our proposed methodology, and how they compare to Prediction Accuracy, AUC and F-measure.

The two types noise addition we use are listed below. Note that these are simple toy noise addition techniques, and are not part of this paper's contribution.

\paragraph{Uniform Noise (UN)} A user-defined percentage of values in the dataset $D$ are changed, with the result being $M$. If a value $r_a$ is changed and $a$ is a continuous attribute, the new value is selected from a uniform distribution between the minimum and maximum values of $a$. If $a$ is a discrete (i.e. categorical) attribute, $r_a$ is changed to any unique value in the set $a$, with each value having an equal probability of being selected. Values are randomly selected, with the original value having no effect on the new value. This has the effect of flattening the distribution of values for all attributes, as well as flattening all multivariate distributions.

\paragraph{Gaussian Noise (GN)} A user-defined percentage of values in the dataset $D$ are changed, with the result being $M$. If a value $r_a$ is changed and $a$ is a continuous attribute, a random number is selected from a Gaussian distribution with a mean of zero and a variance equal to $a$'s variance, and added to $r_a$. If $a$ is a discrete (i.e. categorical) attribute, $r_a$ is changed to a value that is randomly selected from $a$'s original set of values (including repeated values). GN therefore maintains the distribution of values for both numerical and categorical attributes. Additionally, continuous values are changed in a way that takes into account the original value. This means that each record's continuous values are likely to remain close to their original values, and thus the multivariate distribution of the dataset is likely to be preserved.

Neither of these noise types add noise to the label $Y$. For each type of noise, we increase the percentage chance of changing a value in 2\% increments, from 0\% to 30\%. For each increment, for each type of noise, for each dataset, we use 10-fold cross-validation iteratively 10 times (for a total of 100 tests), creating a new set of patterns $Z_D$ each time. For each test, we collect the result of six measures: Pattern Accuracy, PSD, PLD, Prediction Accuracy and AUC and F-measure. 

\subsection{Pearson's Correlation Coefficient}
\label{subsec:pearson}

By comparing the results of each measure as noise increases, for each dataset, we calculate the measures' correlation to each other using Pearson's correlation coefficient (i.e. Pearson's $r$ value) \cite{Pearson1901}. We calculate their correlation for each noise type separately. The coefficient has a range of $-1\leq r\leq1$, where a result close to 1 indicates a high positive correlation (as one measure increases, so does the other measure), a result close to -1 indicates a high negative correlation (as one measure increases, the other decreases), and a result close to 0 indicates low correlation (the result of one measure has little bearing on the result of the other). To standardize the results of different datasets, we look at the \emph{difference} between each measure's result on $M$ compared to $D$. In other words, for each level of noise, we subtract the result that the measure achieved when there was zero noise. This has no effect on our proposed measures (which always equal 0 when there is no noise), and simply causes Prediction Accuracy, AUC and F-measure to be reported as the difference between the ``true'' result and the ``noisy'' result.

\section{Results}
\label{sec:results}

To demonstrate the information a user can learn about individual pattern retention, Fig. \ref{fig:patterns} presents the support and Chi-squared histogram distance of the example patterns shown in Table \ref{table:patterns}, as UN increases. In this example, we can see that the four patterns are affected quite differently by the noise addition. Some of the observations a data scientist could make about these four patterns are:

\begin{figure}%[!t]
%\centering  
\includegraphics[width=1.0\textwidth]{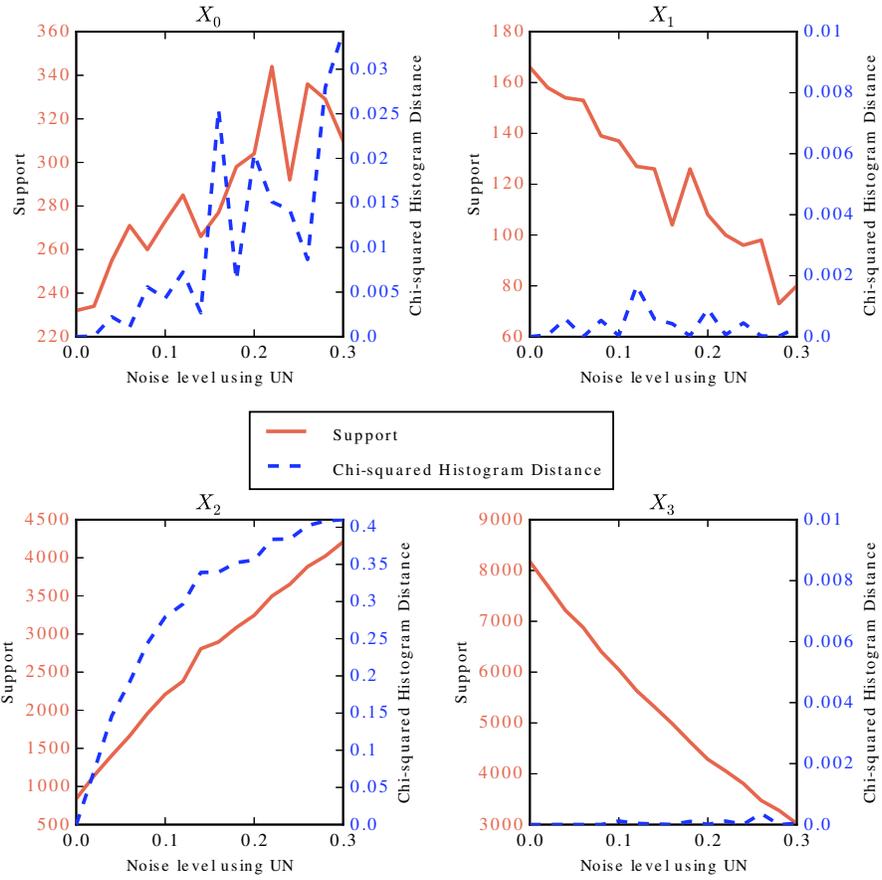}
\caption{The support and Chi-squared Histogram Distance of the example patterns shown in Table \ref{table:patterns} (discovered in the Adult dataset), as UN increases.}
\label{fig:patterns}
\end{figure}

\begin{itemize}
	\item $X_3$ has gone from representing over 8000 of the of 30162 records in Adult to representing only 3000 records in the modified version of Adult by the time UN has reached 30\%.
	\item Despite this massive change in support, the distribution of the class labels in $X_3$ is actually almost exactly the same at all noise levels.
	\item The same cannot be said for $X_2$, where a massive change in support (from less than 1000 to roughly 4000) has been accompanied by a massive change in the distribution of class labels as well.
	\begin{itemize}
		\item If this observation caused the user to investigate further, they would find that $X_2$'s change in label distribution was enough to completely flip the prediction the pattern is making! At 30\% noise, the reported probability of a record having each label is $Pr(Income\leq\$50,000)=0.68$,~$Pr(Income>\$50,000)=0.32$, compared to the probabilities shown in Table \ref{table:patterns}: $Pr(Income\leq\$50,000)=0.05$,~$Pr(Income>\$50,000)=0.95$. It would be incredibly damaging to any analysis performed with $M$ if the data analyst trusted this pattern.
	\end{itemize}
	\item $X_0$ and $X_1$ represent a much smaller proportion of the Adult dataset, and have undergone moderate changes in support. $X_0$ has grown larger, while $X_1$ has become smaller, but neither saw enough change in label distribution to cause concern.
	\item These patterns in Adult were discovered with a decision tree, along with 31 other patterns that underwent a variety of changes in support and label distribution similar to the changes shown in Fig. \ref{fig:patterns}.
\end{itemize}

\begin{figure}%[!t]
%\centering  
\includegraphics[width=1.0\textwidth]{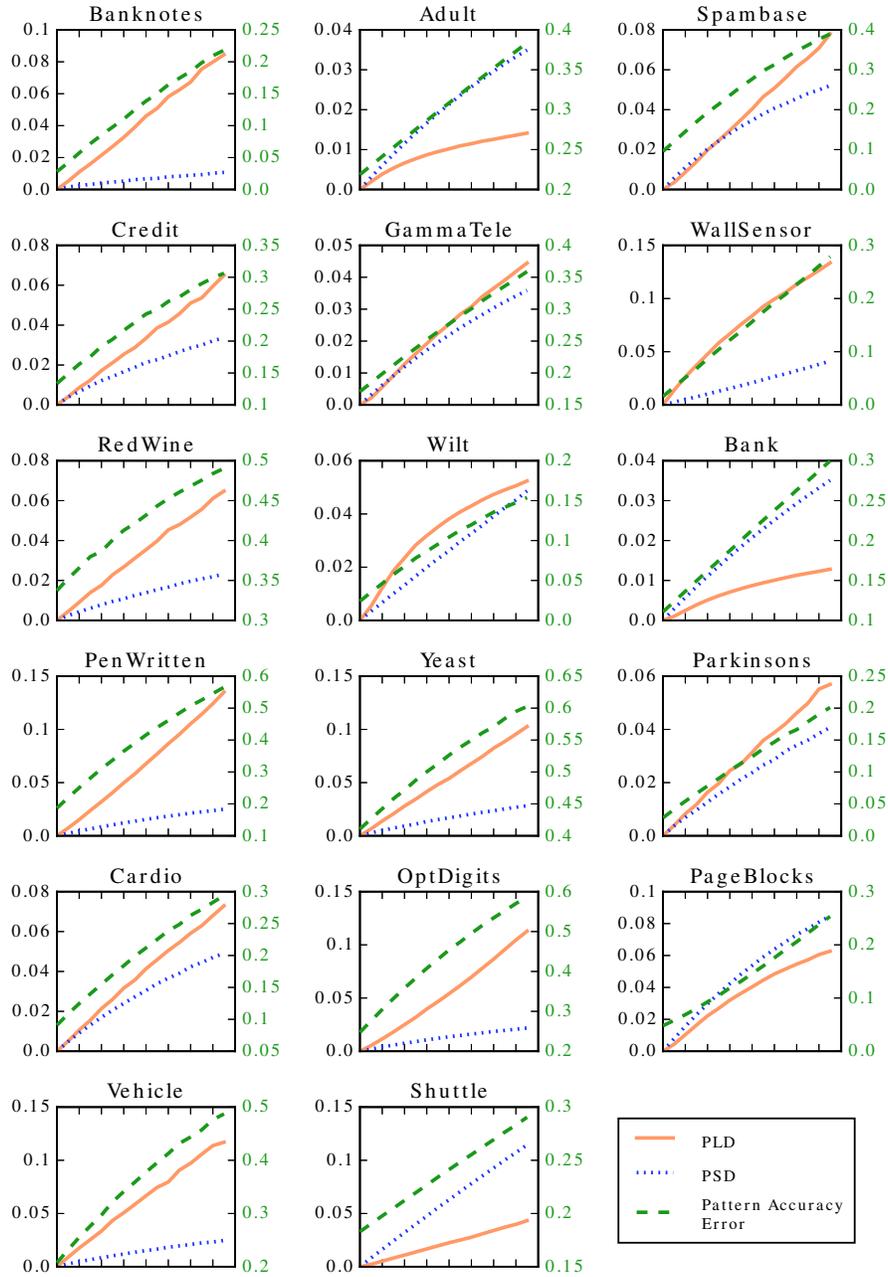}
\caption{The mean results of PSD, PLD and Pattern Accuracy Error as UN increases. The left-hand $y$ axis corresponds to PLD and PSD. The right-hand $y$ axis corresponds to Pattern Accuracy Error. The $x$ axis is the percentage of noise from 0\% to 30\%.}
\label{fig:grid}
\end{figure}

After averaging the support distance and Chi-squared histogram distance of all patterns and thus calculating PSD and PLD, we can compare their assessments of $M$'s pattern retention for each dataset. We also compare PSD and PLD to the assessment made by Pattern Accuracy. For each dataset, we present the results of PSD, PLD and Pattern Accuracy as UN increases in Fig. \ref{fig:grid}. Note that for Pattern Accuracy, we present the percentage of cases where $Z_D$ \emph{incorrectly} predicts the label of records in $M$ so that lower values signify better pattern retention for all three measures.\footnote{That is, $pattern~accuracy~error = 1 - pattern~accuracy$.} As more noise is added, we observe that all three measures trend up as expected, but upon closer inspection we can see that they do not do so at identical rates.

\subsection{Correlations between utility measures}
\label{subsec:correlations}

The differences in trends seen in Fig. \ref{fig:grid} are quantified by the correlations between the measures, seen in Table \ref{table:correlations}. The correlations are calculated using Pearson's correlation coefficient \cite{Pearson1901} as described in Sect. \ref{subsec:pearson}, and Prediction Accuracy, AUC and F-measure are included as well. Unlike Fig. \ref{fig:grid}, the correlations are calculated using the nine datasets with binary labels for the benefit of AUC and F-measure.\footnote{The correlations among PLD, PSD, Pattern Accuracy and Prediction Accuracy when using the datasets shown in Fig. \ref{fig:grid} are similar to those shown in Table \ref{table:correlations}.} 

\begin{table}[!t]
\centering 
\caption{A matrix of correlations for each combination of two measures, for all two noise types. We include the $p$ value of each correlation in brackets (i.e. the probability of observing a result at least as extreme as the one reported by chance, assuming there is zero correlation).}
\scriptsize
	\begin{tabular}{>{\centering\arraybackslash}m{1.5cm}  >{\centering\arraybackslash}m{1.6cm} >{\centering\arraybackslash}m{1.6cm} >{\centering\arraybackslash}m{1.6cm} >{\centering\arraybackslash}m{1.6cm} >{\centering\arraybackslash}m{1.6cm}}
	\noalign{\smallskip}\hline\noalign{\smallskip}
	\textbf{Measure} 		& \textbf{PLD} 		& \textbf{PSD} 		& \textbf{Prediction Accuracy} & \textbf{AUC} 	& \textbf{F-measure} \\
	\noalign{\smallskip}\hline\hline\noalign{\smallskip}
	\multicolumn{6}{c}{\textbf{UN}} \\
	\noalign{\smallskip}\hline\noalign{\smallskip}
	\textbf{Pattern Accuracy}	& -0.77 (0.00)& -0.83 (0.00)		& 0.36 (0.00)		& 0.34 (0.00)		& 0.26 (0.00) \\ \noalign{\smallskip}
	\textbf{PLD}				& ~									& 0.58 (0.00) 		& -0.32 (0.00)	& -0.23 (0.01) 	& -0.09 (0.33) \\ \noalign{\smallskip}
  \textbf{PSD}				& ~									& ~								& -0.21 (0.02) 	& -0.35 (0.00) 	& -0.28 (0.00) \\ \noalign{\smallskip}
	\textbf{Prediction Accuracy}	& ~				& ~								& ~							& 0.77 (0.00)		& 0.47 (0.00) \\ \noalign{\smallskip}
	\textbf{AUC}				& ~									& ~								& ~							& ~							& 0.91 (0.00) \\
	\noalign{\smallskip}\hline\noalign{\smallskip}
	\multicolumn{6}{c}{\textbf{GN}} \\
	\noalign{\smallskip}\hline\noalign{\smallskip}
	\textbf{Pattern Accuracy}	& -0.86 (0.00)& -0.77 (0.00)	& 0.19 (0.04) 			& 0.20 (0.02) 		& 0.21 (0.02) \\ \noalign{\smallskip}
	\textbf{PLD}				& ~									& 0.44 (0.00)		& -0.04 (0.67)			& -0.08 (0.35)		& -0.09 (0.34) \\ \noalign{\smallskip}
  \textbf{PSD}				& ~									& ~							& -0.13 (0.14)			& -0.27 (0.00)		& -0.26 (0.00) \\ \noalign{\smallskip}
	\textbf{Prediction Accuracy}	& ~				& ~							& ~									& 0.72 (0.00)			& 0.50 (0.00) \\ \noalign{\smallskip}
	\textbf{AUC}				& ~									& ~							& ~									& ~								& 0.93 (0.00) \\
	\noalign{\smallskip}\hline
  \end{tabular}
	\label{table:correlations}
\end{table}

One observation we can make about Table \ref{table:correlations} is that despite all the measures using the same data, they do not always agree with each other. Just because Prediction Accuracy decreases does not mean that F-measure also decreases, for example. Another observation is that Prediction Accuracy, F-measure and AUC have very weak correlations with any of our implementations of our proposed methodology. This is interesting, and confirms our suspicions that just because a good classifier (that is, a classifier that achieves good results) can be made from noisy data, does not mean that the patterns in the noisy data have the same properties as the original patterns, or even that the original patterns are in the noisy data at all. For example if a user observed a particular amount of Prediction Accuracy loss after modifying $D$ to $M$, there is no way to tell how much the support of the patterns in $D$ might have changed.

\section{Discussion}
\label{sec:conclusion}

None of our proposed measures can tell a user if a good classifier can be made from $M$. They are not trying to! If a user wishes to learn that, they can use machine learning algorithms on $M$ and see if the resulting classifier has good performance, using measures such as Prediction Accuracy. Doing so, however, will not tell them if those machine learning algorithms found the same patterns that existed in $D$. That is where our proposed methodology -- and our implementations of that methodology -- come in.

Pattern Accuracy, PSD and PLD should not be interpreted as exhaustively measuring all aspects of pattern retention. Rather, they are examples of quantifying specific effects a privacy-preservation technique can have on data. It is the responsibility of the data scientist performing the anonymization of $D$ to assess what properties of a dataset are relevant or important, and then to measure how those properties might have changed after data modification or synthesization. Pattern Accuracy measures the overall retention of the original patterns; PSD measures changes in pattern support, per pattern; PLD measures changes in label distribution per pattern; other measures might focus on quantifying changes in pattern conciseness or peculiarity or any number of other properties that might make patterns interesting to a user.

Prediction Accuracy is currently heavily relied upon in privacy-preservation research. While the measure itself is very useful, it should not be viewed as an all-encompassing measure of the quality of modified or synthetic data, but rather as another example of quantifying a specific property -- the ability for accurate classifiers to be built using a variety of machine-learning algorithms.

Measuring properties of $Z_D$ in $M$ is straightforward, both conceptually and computationally, and can be easily used in conjunction with Prediction Accuracy and other measures. It enables the user to quantify aspects of $M$ that previously could only be assessed with experience or intuition.

\section*{References}

\bibliography{My_Collection}

\begin{thebibliography}{10}
\expandafter\ifx\csname url\endcsname\relax
  \def\url#1{\texttt{#1}}\fi
\expandafter\ifx\csname urlprefix\endcsname\relax\def\urlprefix{URL }\fi
\expandafter\ifx\csname href\endcsname\relax
  \def\href#1#2{#2} \def\path#1{#1}\fi

\bibitem{Breiman2001}
L.~Breiman, {Random forests}, Machine learning (2001) 1--35.

\bibitem{Islam2011a}
M.~Z. Islam, H.~Giggins, {Knowledge discovery through SysFor: a systematically
  developed forest of multiple decision trees}, in: Ninth Australasian Data
  Mining Conference-Volume 121, Australian Computer Society, Inc., Ballarat,
  Australia, 2011, pp. 195--204.

\bibitem{Evfimievski2004}
A.~Evfimievski, R.~Srikant, R.~Agrawal, J.~Gehrke, {Privacy preserving mining
  of association rules}, Information Systems 29~(4) (2004) 343--364.

\bibitem{Han2007}
J.~Han, H.~Cheng, D.~Xin, X.~Yan, {Frequent pattern mining: current status and
  future directions}, Data Mining and Knowledge Discovery 15~(1) (2007) 55--86.

\bibitem{Agrawal2000}
R.~Agrawal, R.~Srikant, {Privacy-preserving Data Mining}, in: Proceedings of
  the 2000 ACM SIGMOD Conference on Management of Data, Dallas, Texas, 2000,
  pp. 439--450.

\bibitem{Agrawal2001}
D.~Agrawal, C.~Aggarwal, {On the design and quantification of privacy
  preserving data mining algorithms}, in: Proceedings of the twentieth ACM
  SIGMOD-SIGACT-SIGART symposium on Principles of database systems., ACM, 2001,
  pp. 247--255.

\bibitem{Islam2011}
M.~Z. Islam, L.~Brankovic, {Privacy preserving data mining: A noise addition
  framework using a novel clustering technique}, Knowledge-Based Systems 24~(8)
  (2011) 1214--1223.

\bibitem{Liu2006}
K.~Liu, H.~Kargupta, J.~Ryan, {Random projection-based multiplicative data
  perturbation for privacy preserving distributed data mining}, Knowledge and
  Data Engineering, IEEE Transactions on 18~(1) (2006) 92--106.

\bibitem{Sweeney2002a}
L.~Sweeney, {k-anonymity: A model for protecting privacy}, International
  Journal of Uncertainty, Fuzziness and Knowledge-Based Systems 10~(05) (2002)
  557--570.

\bibitem{Machanavajjhala2007}
A.~Machanavajjhala, D.~Kifer, J.~Gehrke, M.~Venkitasubramaniam, {l-diversity:
  Privacy beyond k-anonymity}, ACM Transactions on Knowledge Discovery from
  Data (TKDD) 1~(1) (2007) 3.

\bibitem{Dwork2006}
C.~Dwork, {Differential Privacy}, in: Automata, languages and programming, Vol.
  4052, Springer Berlin Heidelberg, Venice, Italy, 2006, pp. 1--12.

\bibitem{Dwork2006a}
C.~Dwork, F.~McSherry, K.~Nissim, A.~Smith, {Calibrating noise to sensitivity
  in private data analysis}, Theory of Cryptography (2006) 265--284.

\bibitem{McSherry2007}
F.~McSherry, K.~Talwar, {Mechanism Design via Differential Privacy}, 48th
  Annual IEEE Symposium on Foundations of Computer Science (FOCS'07) (2007)
  94--103.

\bibitem{Brickell2008}
J.~Brickell, V.~Shmatikov, {The cost of privacy: destruction of data-mining
  utility in anonymized data publishing}, in: Proceedings of the 14th ACM
  SIGKDD international conference on Knowledge discovery and data mining, ACM,
  2008, pp. 70--78.

\bibitem{Fung2005}
B.~Fung, K.~Wang, P.~Yu, {Top-down specialization for information and privacy
  preservation}, in: Proceedings of the 21st International Conference on Data
  Engineering, IEEE, 2005, pp. 205--216.

\bibitem{Han2006a}
J.~Han, M.~Kamber, J.~Pei, {Data mining: concepts and techniques}, Morgan
  Kaufmann Publishers, 2006.

\bibitem{Rijsbergen1979}
C.~van Rijsbergen, {Information Retrieval}, Butterworth, 1979.

\bibitem{Hanley1982}
J.~Hanley, B.~McNeil, {The meaning and use of the area under a receiver
  operating characteristic (ROC) curve}, Radiology 143~(1) (1982) 29--36.

\bibitem{Bache2013}
K.~Bache, M.~Lichman, \href{http://archive.ics.uci.edu/ml/}{{UCI Machine
  Learning Repository}} (2013).
\newline\urlprefix\url{http://archive.ics.uci.edu/ml/}

\bibitem{Tan2002}
P.-N. Tan, V.~Kumar, J.~Srivastava, {Selecting the right interestingness
  measure for association patterns}, in: Proceedings of the eighth ACM SIGKDD
  international conference on Knowledge discovery and data mining, Vol.~2, ACM
  Press, New York, USA, 2002, p.~32.

\bibitem{Webb2002}
G.~Webb, D.~Brain, {Generality is predictive of prediction accuracy}, in:
  Proceedings of the 2002 Pacific Rim Knowledge Acquisition Workshop (PKAW
  2002), 2002, pp. 117--130.

\bibitem{Dasseni2001}
E.~Dasseni, V.~Verykios, A.~K. Elmagarmid, E.~Bertino, {Hiding Association
  Rules by Using Confidence and Support}, in: Information Hiding, Purdue
  University, Springer Berlin Heidelberg, 2001, pp. 369--383.

\bibitem{Padmanabhan2000}
B.~Padmanabhan, A.~Tuzhilin, {Small is beautiful: discovering the minimal set
  of unexpected patterns}, in: Proceedings of the 6th ACM SIGKDD international
  conference on Knowledge discovery and data mining, ACM, 2000, pp. 54--63.

\bibitem{Ohshima2003}
N.~Zhong, Y.~Yao, M.~Ohshima, {Peculiarity oriented multidatabase mining}, IEEE
  Transactions on Knowledge and Data Engineering 15~(4) (2003) 952--960.

\bibitem{Geng2006}
L.~Geng, H.~J. Hamilton, {Interestingness measures for data mining: a survey},
  ACM Computing Surveys 38~(3) (2006) 9.

\bibitem{Tan2004}
P.-N. Tan, V.~Kumar, J.~Srivastava, {Selecting the right objective measure for
  association analysis}, Information Systems 29~(4) (2004) 293--313.

\bibitem{Vaillant2004}
B.~Vaillant, P.~Lenca, S.~Lallich, {A clustering of interestingness measures},
  in: Discovery Science, Springer Berlin Heidelberg, 2004, pp. 290--297.

\bibitem{Fletcher2015}
S.~Fletcher, M.~Z. Islam, {Measuring Information Quality for Privacy Preserving
  Data Mining}, International Journal of Computer Theory and Engineering 7~(1)
  (2015) 21--28.

\bibitem{Breiman1984}
L.~Breiman, J.~Friedman, C.~Stone, R.~Olshen, {Classification and regression
  trees}, Chapman \& Hall/CRC, 1984.

\bibitem{Ferri2009}
C.~Ferri, J.~Hern\'{a}ndez-Orallo, R.~Modroiu, {An experimental comparison of
  performance measures for classification}, Pattern Recognition Letters 30~(1)
  (2009) 27--38.

\bibitem{Sokolova2009}
M.~Sokolova, G.~Lapalme, {A systematic analysis of performance measures for
  classification tasks}, Information Processing \& Management 45~(4) (2009)
  427--437.

\bibitem{Fletcher2014}
S.~Fletcher, M.~Z. Islam, {Quality evaluation of an anonymized dataset}, in:
  22nd International Conference on Pattern Recognition, IEEE, Stockholm,
  Sweden, 2014, pp. 3594--3599.

\bibitem{Estivil-Castro1999}
V.~Estivill-Castro, L.~Brankovic, {Data Swapping: Balancing Privacy Against
  Precision in Mining for Logic Rules}, Data Warehousing and Knowledge
  Discovery (DaWaK '99) (1999) 389--398.

\bibitem{Pele2010}
O.~Pele, M.~Werman, {The quadratic-chi histogram distance family}, Lecture
  Notes in Computer Science (including subseries Lecture Notes in Artificial
  Intelligence and Lecture Notes in Bioinformatics) 6312 LNCS~(PART 2) (2010)
  749--762.

\bibitem{Felkin2007}
M.~Felkin, {Comparing classification results between n-ary and binary
  problems}, in: Quality Measures in Data Mining, Springer Berlin Heidelberg,
  2007, Ch.~12, pp. 277--301.

\bibitem{Hand2009}
D.~J. Hand, {Measuring classifier performance: a coherent alternative to the
  area under the ROC curve}, Machine Learning 77~(1) (2009) 103--123.

\bibitem{Lobo2008}
J.~M. Lobo, A.~Jim\'{e}nez-valverde, R.~Real, {AUC: A misleading measure of the
  performance of predictive distribution models}, Global Ecology and
  Biogeography 17~(2) (2008) 145--151.

\bibitem{Pearson1901}
K.~Pearson, {On lines and planes of closest fit to systems of points in space},
  The London, Edinburgh, and Dublin Philosophical Magazine and Journal of
  Science 2~(11) (1901) 559--572.

\end{thebibliography}

\end{document}